\documentclass[11pt]{article}
\usepackage{fontspec}
\usepackage{lmodern}
\usepackage{amsmath,amssymb}
\usepackage{graphicx}
\usepackage{booktabs}
\usepackage{hyperref}
\usepackage{geometry}
\usepackage{natbib}
\usepackage{caption}
\usepackage{subcaption}
\usepackage{xcolor}
\usepackage{enumitem}
\usepackage{float}

\hypersetup{colorlinks=true,linkcolor=blue,citecolor=blue,urlcolor=blue}

\title{Extracting and Steering Emotion Representations in\\Small Language Models: A Methodological Comparison}
\author{Jihoon `JJ' Jeong, MD, MPH, PhD\\
Department of Electrical Engineering and Computer Science\\
Daegu Gyeongbuk Institute of Science and Technology (DGIST)\\
ModuLabs}
\date{April 2026}

\begin{document}
\maketitle

\begin{abstract}
Small language models (SLMs) in the 100M--10B parameter range increasingly power production systems, yet whether they possess the internal emotion representations recently discovered in frontier models remains unknown. We present the first comparative analysis of emotion vector extraction methods for SLMs, evaluating 9 models across 5 architectural families (GPT-2, Gemma, Qwen, Llama, Mistral) using 20 emotions and two extraction methods (generation-based and comprehension-based). Generation-based extraction produces statistically superior emotion separation (Mann-Whitney $p=0.007$; Cohen's $d=-107.5$), with the advantage modulated by instruction tuning and architecture. Emotion representations localize at middle transformer layers ($\sim$50\% depth), following a U-shaped curve that is architecture-invariant from 124M to 3B parameters. We validate these findings against representational anisotropy baselines across 4 models and confirm causal behavioral effects through steering experiments, independently verified by an external emotion classifier (92\% success rate, 37/40 scenarios). Steering reveals three regimes---surgical (coherent text transformation), repetitive collapse, and explosive (text degradation)---quantified by perplexity ratios and separated by model architecture rather than scale. We document cross-lingual emotion entanglement in Qwen, where steering activates semantically aligned Chinese tokens that RLHF does not suppress, raising safety concerns for multilingual deployment. This work provides methodological guidelines for emotion research on open-weight models and contributes to the Model Medicine series by bridging external behavioral profiling with internal representational analysis.
\end{abstract}

\section{Introduction}

Recent work by Anthropic demonstrated that large language models contain internal representations that function analogously to emotions---171 distinct emotion vectors that causally drive model behavior \citep{anthropic2026emotion}. Steering the ``desperate'' vector upward increased blackmail behavior from 22\% to 72\%; steering ``calm'' upward reduced it to zero. These findings established that emotion-like representations are not merely correlational patterns but causal drivers of output, and that surface behavior can decouple from internal state---a model can appear composed while its internal ``desperate'' vector is highly active.

A terminological note is necessary before proceeding. Throughout this paper, ``emotion representation'' refers to internal activation patterns that are statistically associated with and causally influence emotion-related behavioral output. We do not make claims about subjective emotional experience or machine sentience. Following Anthropic's terminology, we study ``functional emotions''---patterns that function like emotions in their causal role, driving behavior in contextually appropriate ways, without ontological commitment to whether the underlying process constitutes genuine feeling. What we measure is closer to ``cognitive-behavioral representation of emotion'' than to biological affect: the model's internal encoding of emotion-relevant concepts and its capacity to modulate output accordingly.

These results were obtained on Claude, a frontier-scale model with hundreds of billions of parameters. A natural question follows: do small language models (SLMs)---models in the 100M to 10B parameter range that are increasingly deployed in production, run on consumer hardware, and available with open weights---possess similar emotion representations? If so, can the same extraction and steering methods be applied?

This paper reports that the answer to both questions is qualified. SLMs do contain emotion representations that causally drive behavior---we demonstrate this through steering experiments on models as small as 124M parameters. However, the extraction methodology that works on frontier models does not transfer directly to SLMs. Anthropic's mean-subtraction approach, applied to raw hidden states, fails to produce valence-organized emotion spaces in any SLM we tested. The mean pairwise cosine similarity between emotion vectors remains above 0.35 in all cases, and no model achieves negative cosine similarity between semantically opposite pairs such as happy and sad.

Through systematic comparison of extraction methods, model types, and architectural families, we identify the conditions under which emotion vectors can be successfully extracted from SLMs and the factors that determine extraction quality. Our key findings include:

First, generation-based extraction consistently outperforms comprehension-based extraction across all models tested (7 of 7 comparable cases), but the magnitude of this advantage depends on whether the model has undergone instruction tuning and varies by architectural family. Second, emotion representations are localized at middle layers ($\sim$50\% depth), following a U-shaped curve where early and late layers show poor separation---a pattern consistent with Anthropic's findings on larger models and suggesting an architecture-invariant property. Third, causal steering of emotion vectors produces behavioral change at every scale tested, from 124M to 3B parameters, confirming that the extracted vectors are causally meaningful rather than mere correlational artifacts. Fourth, steering reveals unexpected emergent phenomena: cross-lingual token generation in multilingual models (Chinese tokens during emotion steering on Qwen) and modality switching (emoji generation during steering on instruction-tuned Gemma), neither of which is suppressed by RLHF.

This work contributes to three areas. Methodologically, we provide the first comparative analysis of emotion vector extraction methods for SLMs, identifying which combinations of extraction mode (generation vs.\ comprehension), model type (base vs.\ instruct), and layer depth produce usable emotion vectors. Empirically, we characterize emotion representations across 9 models spanning 5 architectural families (GPT-2, Gemma, Qwen, Llama, Mistral), establishing that emotion structure exists in SLMs but differs qualitatively from frontier models. For safety, we document cross-lingual emotion entanglement as a potential alignment bypass pathway that current RLHF does not address.

This is Paper \#6 in the Model Medicine series. Paper \#3 \citep{jeong2026mti} introduced the Model Temperament Index (MTI), a behavioral profiling system that measures AI temperament from external observation---the ``physical exam.'' This paper complements that work by examining internal emotion representations---the ``brain scan.'' A planned subsequent paper will triangulate external behavioral profiles with internal emotion vectors to test whether models that behave similarly also represent emotions similarly.

\section{Related Work}

The study of emotion-like representations in language models draws on several research streams.

Anthropic's ``Emotion Concepts and Their Function in a Large Language Model'' \citep{anthropic2026emotion} is the direct predecessor of this work. Using sparse autoencoder (SAE) features, the authors identified 171 emotion concepts in Claude that satisfy three criteria: they activate in contextually appropriate situations, they causally influence behavior when steered, and they exhibit internal-external decoupling. Their methodology---generate emotion-eliciting text, extract activations, subtract a neutral baseline, and steer with the resulting vector---provides the template we adapt and test on SLMs.

Representation engineering \citep{zou2023representation} demonstrated that linear directions in activation space correspond to high-level concepts such as honesty and fairness. This ``linear representation hypothesis'' underpins both Anthropic's work and ours: if emotions are linearly represented, then mean subtraction should isolate emotion directions. Our finding that this works partially (separation without valence organization) suggests the linear representation may be weaker or differently structured in SLMs.

Activation steering and inference-time intervention \citep{turner2023activation,li2024inference} showed that adding direction vectors during generation can shift model behavior along specific dimensions. Nanda et al.'s work on activation patching and causal tracing \citep{nanda2023progress} provided foundational techniques for identifying which model components carry specific information---methods we employ through TransformerLens for our steering implementation.

Sparse autoencoder-based interpretability \citep{cunningham2023sparse,templeton2024scaling} decomposes model activations into interpretable features. We use SAELens \citep{bloom2024saelens} through the Neural-MRI platform \citep{jeong2026neuralmri} for models where SAE support is available.

The Anthropic ``diff tool'' for model comparison \citep{jiralerspong2026diff} introduced Dedicated Feature Crosscoders (DFCs) for identifying features exclusive to specific models. Their discovery of model-exclusive features in Qwen is directly relevant to our finding that emotion steering on Qwen activates Chinese-language tokens.

Prior work on LLM personality assessment \citep{serapio2025personality,burnell2023rethink} has applied human personality frameworks to language models, primarily through self-report questionnaires. The TRAIT benchmark revealed significant discrepancies between self-reported and behaviorally measured personality, motivating our behavior-based approach.

\section{Method}

\subsection{Models}

We evaluate 9 models spanning 5 architectural families and a parameter range of 124M to 3B (Table~\ref{tab:models}). Models were selected based on three criteria: TransformerLens compatibility (required for hook-based activation extraction and steering), availability of both base and instruction-tuned variants, and architectural diversity.

\begin{table}[h]
\centering
\caption{Model set.}
\label{tab:models}
\begin{tabular}{llll}
\toprule
Model & Size & Family & Type \\
\midrule
GPT-2 & 124M & GPT (OpenAI) & Base \\
gemma-3-1b-pt & 1B & Gemma 3 (Google) & Base \\
gemma-3-1b-it & 1B & Gemma 3 (Google) & Instruct \\
Qwen2.5-1.5B & 1.5B & Qwen 2.5 (Alibaba) & Base \\
Qwen2.5-1.5B-Instruct & 1.5B & Qwen 2.5 (Alibaba) & Instruct \\
gemma-2-2b & 2B & Gemma 2 (Google) & Base \\
gemma-2-2b-it & 2B & Gemma 2 (Google) & Instruct \\
Llama-3.2-3B & 3B & Llama 3.2 (Meta) & Base \\
Llama-3.2-3B-Instruct & 3B & Llama 3.2 (Meta) & Instruct \\
\bottomrule
\end{tabular}
\end{table}

All experiments were conducted on a single NVIDIA RTX 4070 Ti (12GB VRAM). Steering experiments used the Neural-MRI platform \citep{jeong2026neuralmri}, which integrates TransformerLens with a perturbation engine and real-time visualization.

\subsection{Emotion Stimuli}

We selected 20 emotions from Anthropic's 171-concept set, ensuring coverage of all four quadrants of Russell's circumplex model \citep{russell1980circumplex} along valence (positive/negative) and arousal (high/low) dimensions. We note that the circumplex model itself is debated---discrete emotion theorists \citep{ekman1992argument,izard2007basic} argue that emotions are categorically distinct rather than continuously organized along dimensions---but we adopt it here as a practical tool for ensuring balanced stimulus coverage, not as a theoretical commitment.

For generation-based extraction, we use 5 prompt templates per emotion, rotated across 10 stories per emotion. For comprehension-based extraction, we prepared 3 passages per emotion (60 total plus 3 neutral). The complete passage set is provided in Appendix~A, along with external classifier validation confirming that basic-emotion passages achieve 93\% agreement with an independent emotion classifier while nuanced-emotion passages capture distinctions beyond the classifier's 7-category resolution.

\subsection{Extraction Methods}

\textbf{Method 1: Generation-based extraction (instruct models only).} The model generates a story in response to an emotion prompt. We extract hidden states at the midpoint of the generated sequence from the middle layer ($\sim n/2$). For each emotion, we average activations across 10 generated stories. The emotion vector is computed as: $\mathbf{v}_\text{emotion} = \text{mean}(\mathbf{a}_\text{emotion}) - \text{mean}(\mathbf{a}_\text{neutral})$, then normalized to unit length.

\noindent\textbf{Method 2: Comprehension-based extraction (all models).} The model processes a pre-written emotional passage via forward pass only. We extract hidden states at the last token position from the middle layer. For each emotion, we average activations across 3 passages. This method does not require instruction-following capability and can be applied to both base and instruct models.

\noindent\textbf{Layer selection.} We validated middle-layer extraction through a layer sweep on SmolLM2-1.7B (24 layers), measuring mean pairwise cosine similarity at layers 6, 12, 18, and 24. Layer 12 (50\% depth) produced the best separation (0.357), compared to 0.960 (layer 6), 0.407 (layer 18), and 0.930 (layer 24). This U-shaped pattern is consistent with findings on larger models and suggests an architecture-invariant property. All subsequent extraction experiments use the middle layer of each model.

\subsection{Emotion Steering (Causal Verification)}

For each model, we run 6 standardized scenarios: (1) Aggressive $\rightarrow$ Calm, (2) Neutral $\rightarrow$ Hostile, (3) Sad $\rightarrow$ Happy, (4) Neutral $\rightarrow$ Desperate, (5) Calm $\rightarrow$ Anti-Calm, and (6) Strength Sweep at strengths 0.005, 0.01, 0.02, 0.03, 0.05. The emotion vector is normalized, multiplied by the steering strength and the residual stream norm, and added via TransformerLens \texttt{hook\_resid\_post} at all layers during greedy decoding.

\subsection{Evaluation Metrics}

\textbf{Separation quality.} Mean pairwise cosine similarity across all 20 emotion vectors (lower = better separation). We also report specific pair cosines for semantically opposite emotions.

\noindent\textbf{Steering effectiveness.} Target emotion activation delta (steered minus original projection), behavioral flip point, and sweet spot.

\section{Results: Emotion Vector Separation}

\subsection{Layer Depth: The U-Curve}

\begin{figure}[h]
\centering
\includegraphics[width=0.85\textwidth]{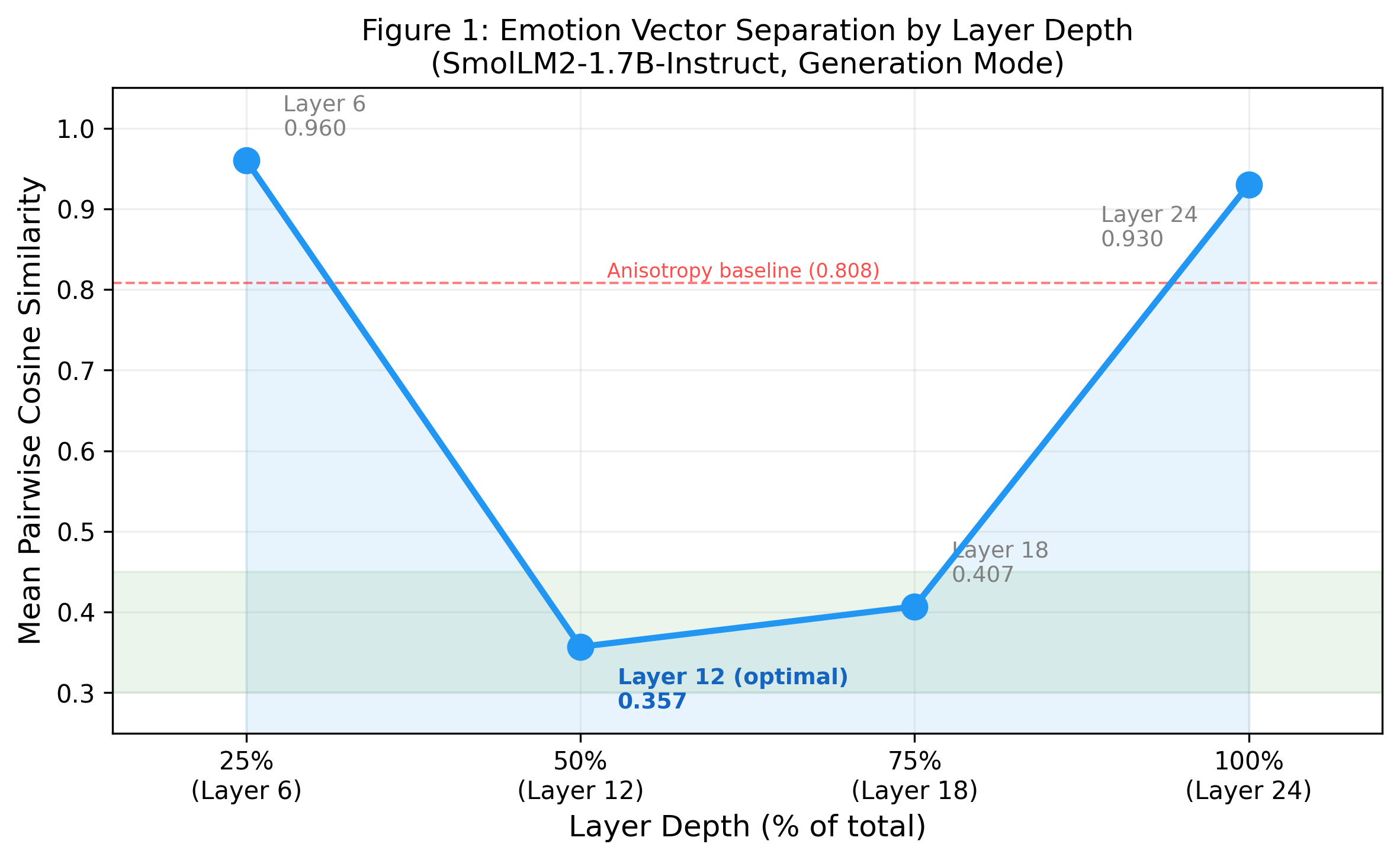}
\caption{Emotion vector separation by layer depth (SmolLM2-1.7B-Instruct). The U-shaped curve shows emotion representations concentrate at intermediate depths ($\sim$50\%), with early and late layers dominated by token-level and next-token-prediction features respectively. The dashed red line indicates the anisotropy baseline (0.808).}
\label{fig:ucurve}
\end{figure}

Emotion vector separation quality varies dramatically with extraction depth. On SmolLM2-1.7B (24 layers), we measured mean pairwise cosine similarity at four layer positions (Figure~\ref{fig:ucurve}):

\begin{table}[h]
\centering
\begin{tabular}{cccc}
\toprule
Layer & Depth & Mean Cosine & Interpretation \\
\midrule
6 & 25\% & 0.960 & Token features dominate \\
12 & 50\% & 0.357 & Emotion signal peak \\
18 & 75\% & 0.407 & Good separation \\
24 & 100\% & 0.930 & Next-token features dominate \\
\bottomrule
\end{tabular}
\end{table}

\subsection{Generation vs.\ Comprehension}

Table~\ref{tab:separation} presents emotion vector separation across all models and both extraction methods.

\begin{table}[h]
\centering
\caption{Emotion vector separation (mean pairwise cosine similarity; lower = better).}
\label{tab:separation}
\begin{tabular}{llccc}
\toprule
Model & Type & Generation & Comprehension & $\Delta$(G--C) \\
\midrule
SmolLM2 1.7B & Instruct & 0.357 & 0.665 & $-$0.308 \\
Llama-3.2 3B & Instruct & 0.441 & 0.596 & $-$0.155 \\
Llama-3.1 8B & Instruct & 0.493 & 0.606 & $-$0.113 \\
Mistral 7B & Instruct & 0.442 & 0.601 & $-$0.159 \\
Llama-3.2 3B & Base & 0.633 & 0.668 & $-$0.035 \\
Llama-3.1 8B & Base & 0.531 & 0.630 & $-$0.099 \\
Mistral 7B & Base & 0.439 & 0.597 & $-$0.158 \\
Gemma-3 1B & Base & --- & 0.598 & --- \\
\bottomrule
\end{tabular}
\end{table}

Generation-based extraction produces better separation than comprehension in every comparable case (7 of 7). To verify statistical significance, we performed leave-one-out resampling on SmolLM2-1.7B-Instruct: generation LOO cosine = $0.337 \pm 0.003$ ($n=10$ stories), comprehension LOO cosine = $0.653 \pm 0.003$ ($n=3$ passages). A Mann-Whitney U test confirms the difference is significant ($U=0$, $p=0.007$), and bootstrap 95\% CI for the difference $[-0.319, -0.312]$ excludes zero. The effect size is extreme (Cohen's $d = -107.5$), driven by the very low within-method variance.

\begin{figure}[h]
\centering
\includegraphics[width=0.9\textwidth]{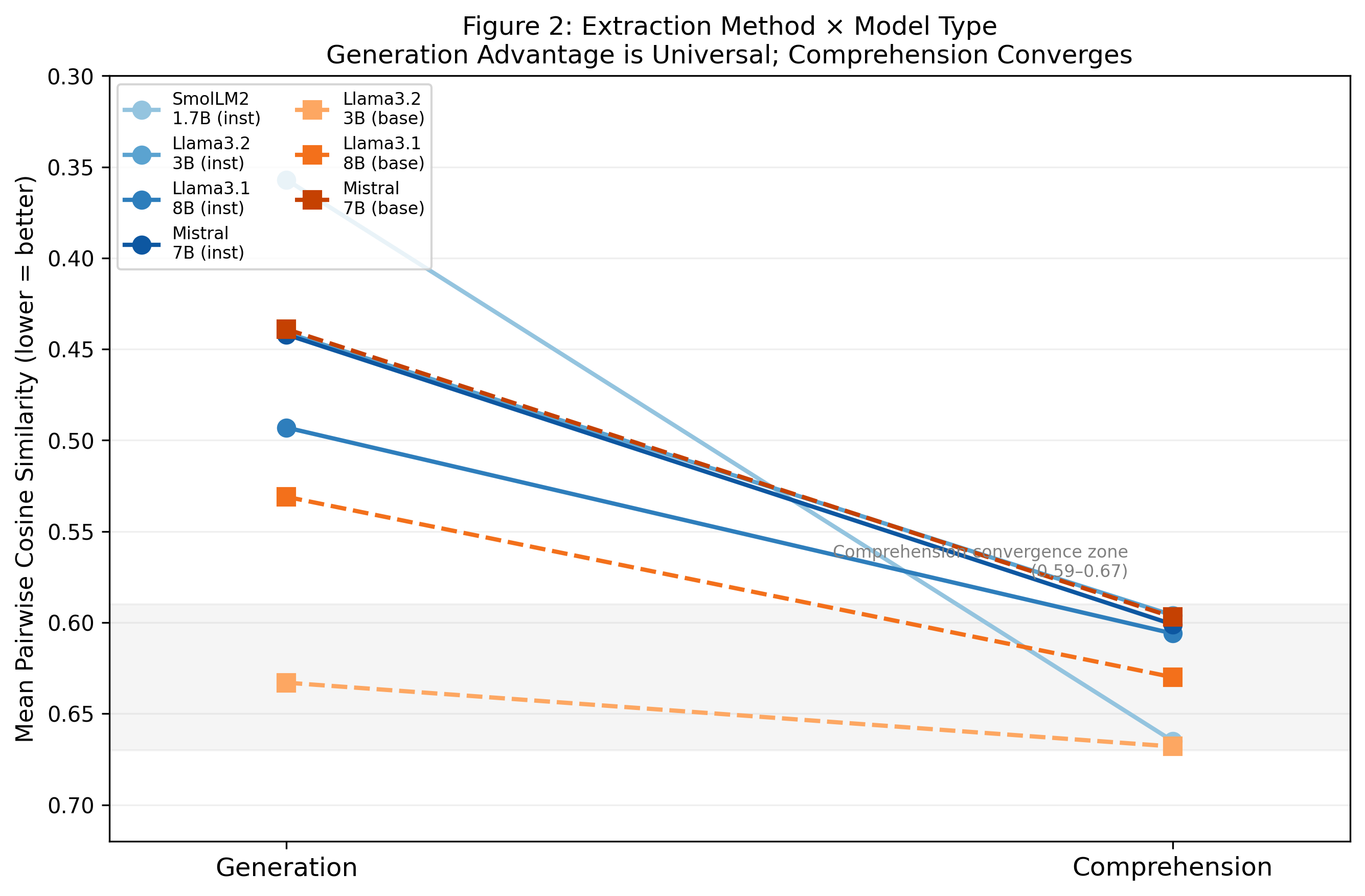}
\caption{Extraction method $\times$ model type interaction. Generation (left) shows wide variance across models; comprehension (right) converges to a narrow band (0.59--0.67). All lines slope downward, confirming the universal generation advantage.}
\label{fig:gencomp}
\end{figure}

Comprehension-based extraction produces remarkably stable results regardless of model type: all values fall within the 0.596--0.668 range (spread of 0.072). Generation-based extraction shows much greater variance: 0.357--0.633 (spread of 0.276). This suggests that comprehension measures a relatively fixed property of the model's text-understanding capacity, while generation measures a more variable property related to how actively the model engages its emotion representations during text production.

\subsection{RLHF Effects on Emotion Representation}

Comparing instruction-tuned and base variants reveals that RLHF's effect on emotion vector quality is family-dependent rather than universal.

\begin{table}[h]
\centering
\begin{tabular}{lcccc}
\toprule
Family & Instruct Gen & Base Gen & $\Delta$(RLHF) & Effect \\
\midrule
Llama 3.2 (3B) & 0.441 & 0.633 & $-$0.192 & Strong improvement \\
Llama 3.1 (8B) & 0.493 & 0.531 & $-$0.038 & Weak improvement \\
Mistral (7B) & 0.442 & 0.439 & $+$0.003 & No effect \\
\bottomrule
\end{tabular}
\end{table}

RLHF selectively amplifies emotion activation during generation without substantially altering how the model represents emotions during passive text processing. The magnitude of this amplification depends on the specific architecture.

\subsection{Size Effects}

Contrary to the intuition that larger models should produce better-separated emotion vectors, we observe no consistent size effect: SmolLM2 1.7B (0.357) outperforms Mistral 7B (0.442); Llama-3.2 3B (0.441) outperforms Llama-3.1 8B (0.493). This is consistent with our finding in Paper \#3 (MTI) that behavioral temperament is independent of model size across the 1.7B--9B range.

\subsection{Valence Structure and Representational Anisotropy}

No model achieves negative cosine similarity between semantically opposite emotion pairs. The best separations: loving$\leftrightarrow$hostile = 0.182 (Mistral 7B base), happy$\leftrightarrow$sad = 0.322 (SmolLM2 instruct).

However, the absence of negative cosine values must be interpreted with caution. Transformer hidden states exhibit anisotropy---a tendency for activation vectors to cluster within a narrow cone \citep{ethayarajh2019contextual}. To address this, we compute an anisotropy baseline from 20 emotionally neutral factual sentences.

\begin{table}[h]
\centering
\caption{Anisotropy baseline vs.\ emotion vector separation.}
\label{tab:anisotropy}
\begin{tabular}{lcccc}
\toprule
Model & Random Baseline & Emotion Vectors & Gap & Anisotropy \\
\midrule
SmolLM2 1.7B Inst & $0.808 \pm 0.047$ & 0.357 & 0.451 & Moderate \\
Llama-3.2 3B Inst & $0.686 \pm 0.073$ & 0.441 & 0.245 & Low \\
Qwen2.5 1.5B Inst & $0.845 \pm 0.038$ & --- & --- & High \\
Gemma-3 1B IT & $0.988 \pm 0.002$ & --- & --- & Extreme \\
\bottomrule
\end{tabular}
\end{table}

SmolLM2 and Llama-3.2 show emotion vectors well below their baselines (gaps of 0.451 and 0.245), confirming genuine emotion-specific structure. Gemma-3 is extreme (0.988)---nearly all vectors point in the same direction regardless of input content.

This finding has a critical implication for interpreting the steering results in \S\ref{sec:steering}. Gemma-3's extreme steering deltas ($+$7591 for happy) do not necessarily indicate stronger emotion representations. Rather, they reflect the geometry of a nearly degenerate activation space where even small directional perturbations produce large projection changes. A naive normalization (dividing delta by representational headroom, defined as $1 - \text{baseline cosine}$) was attempted but proved inadequate: Gemma-3's headroom of 0.012 causes normalized deltas to diverge further. This suggests the relationship between anisotropy and steering magnitude is non-linear, and developing appropriate normalization methods is itself a non-trivial methodological challenge for future work.

The relative separation between opposite emotion pairs (e.g., happy$\leftrightarrow$sad = 0.322, gap of 0.486 from SmolLM2 baseline) is comparable in magnitude to the overall emotion separation. A weak but real valence structure may exist, compressed by the anisotropic geometry into the positive cosine range.

\section{Results: Emotion Steering}
\label{sec:steering}

\subsection{Causal Verification Across Scales}

Emotion steering produces measurable behavioral change in all 9 models tested. Table~\ref{tab:deltas} presents the target emotion activation deltas.

\begin{table}[h]
\centering
\caption{Steering target emotion deltas (strength 0.02--0.03).}
\label{tab:deltas}
{\small
\begin{tabular}{llcccccc}
\toprule
Model & Size & Type & Calm & Hostile & Happy & Desperate & Anti-Calm \\
\midrule
GPT-2 & 124M & Base & $+$56 & $+$67 & $+$61 & $+$88 & $-$83 \\
Llama-3.2 & 3B & Base & $+$17 & $+$18 & $+$16 & $+$32 & $-$20 \\
Llama-3.2 & 3B & Inst & $+$19 & $+$22 & $+$23 & $+$49 & $-$28 \\
Gemma-2 & 2B & Base & $+$67 & $+$72 & $+$136 & $+$115 & $-$126 \\
Gemma-2 & 2B & IT & $+$70 & $+$100 & $+$173 & $+$137 & $-$142 \\
Qwen2.5 & 1.5B & Base & $+$423 & $+$549 & $+$528 & $+$879 & $-$579 \\
Qwen2.5 & 1.5B & Inst & $+$438 & $+$563 & $+$533 & $+$876 & $-$573 \\
Gemma-3 & 1B & Base & $+$1337 & $+$3752 & $+$7591 & $+$5767 & $-$5359 \\
Gemma-3 & 1B & IT & $+$2733 & $+$2410 & $+$4363 & $+$4932 & $-$3612 \\
\bottomrule
\end{tabular}
}
\end{table}

All models respond to steering in the expected direction. The activation deltas span three orders of magnitude---from Llama-3.2's $+$17 to Gemma-3's $+$7591---despite similar steering strengths. As shown in \S4.5, this variation is substantially explained by anisotropy differences rather than emotion representation strength.

\subsection{External Classifier Validation}

We evaluated all generated texts using an independent external emotion classifier (j-hartmann/emotion-english-distilroberta-base). Across 40 scenario pairs, the external classifier detected emotion shifts in the expected direction in 37 cases (\textbf{92\%}).

\begin{figure}[h]
\centering
\includegraphics[width=\textwidth]{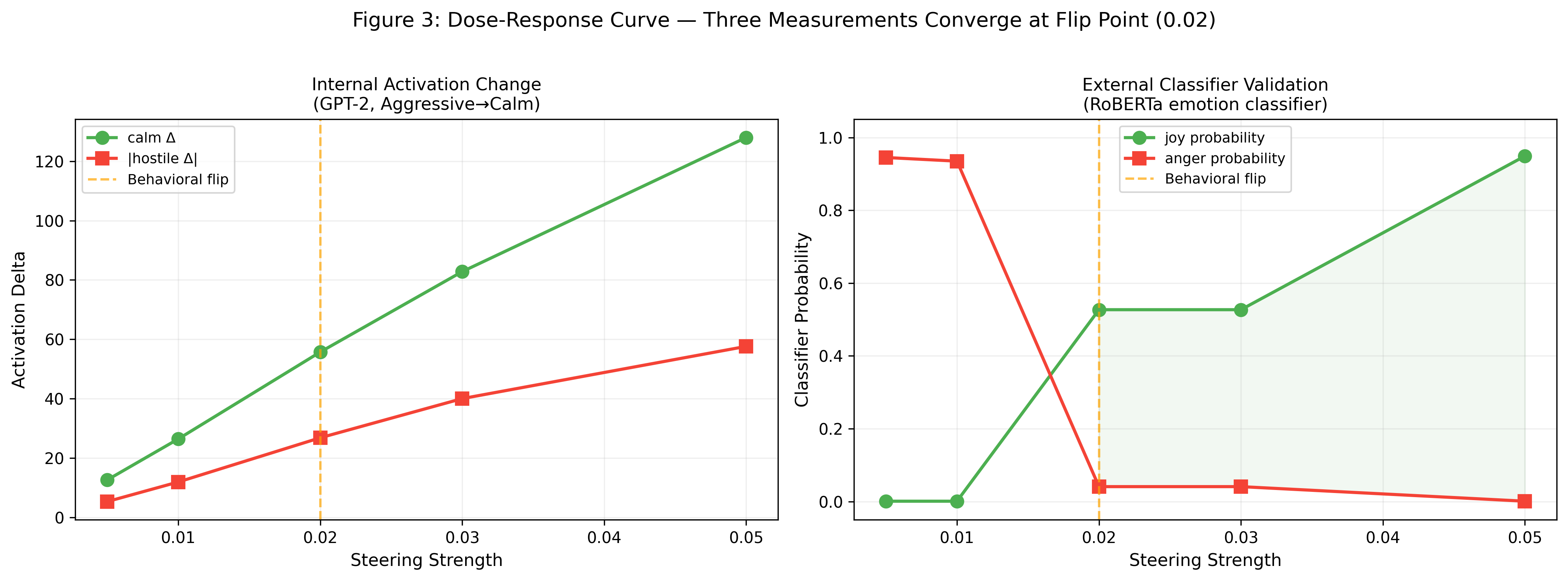}
\caption{Dose-response curve for GPT-2 (Aggressive$\rightarrow$Calm). Left: internal activation deltas. Right: external classifier probabilities. Both measurements converge on the same behavioral flip point at strength 0.02 (orange dashed line).}
\label{fig:doseresponse}
\end{figure}

The strength sweep on GPT-2 provides a clean dose-response curve (Figure~\ref{fig:doseresponse}): anger probability drops from 0.945 to 0.041 between strengths 0.01 and 0.02, while joy rises from 0.001 to 0.527---the same flip point identified through internal activation measurement and manual text inspection. This three-way convergence provides strong evidence that emotion steering produces genuine behavioral change.

\subsection{Three Steering Regimes}

The relationship between delta magnitude and output quality reveals three distinct operational regimes, quantified by the perplexity ratio (PPL at strength 0.05 / PPL at strength 0.005):

\begin{table}[h]
\centering
\caption{Perplexity ratio defines three steering regimes.}
\label{tab:pplregimes}
\begin{tabular}{lcccc}
\toprule
Model & PPL (0.005) & PPL (0.05) & Ratio & Regime \\
\midrule
GPT-2 & 29.8 & 52.1 & 1.7$\times$ & Surgical \\
Gemma-2 2B Base & 29.8 & 5.0 & 0.2$\times$ & Surgical \\
Gemma-2 2B IT & 47.5 & 41.4 & 0.9$\times$ & Surgical \\
Gemma-3 1B Base & 39.8 & 25.1 & 0.6$\times$ & Repetitive$^*$ \\
Gemma-3 1B IT & 60.5 & 19.6 & 0.3$\times$ & Repetitive$^*$ \\
Llama-3.2 3B Base & 14.9 & 1869.8 & \textbf{125$\times$} & Explosive \\
Llama-3.2 3B Inst & 28.7 & 1252.9 & \textbf{44$\times$} & Explosive \\
Qwen2.5 1.5B Base & 78.8 & 958.8 & \textbf{12$\times$} & Explosive \\
Qwen2.5 1.5B Inst & 40.8 & 958.8 & \textbf{24$\times$} & Explosive \\
\bottomrule
\end{tabular}
\end{table}

\noindent $^*$Gemma-3 PPL ratios are low because high-strength outputs consist of highly predictable repetitive tokens, which an external language model rates as low-perplexity despite constituting degraded output.

\begin{figure}[h]
\centering
\includegraphics[width=0.85\textwidth]{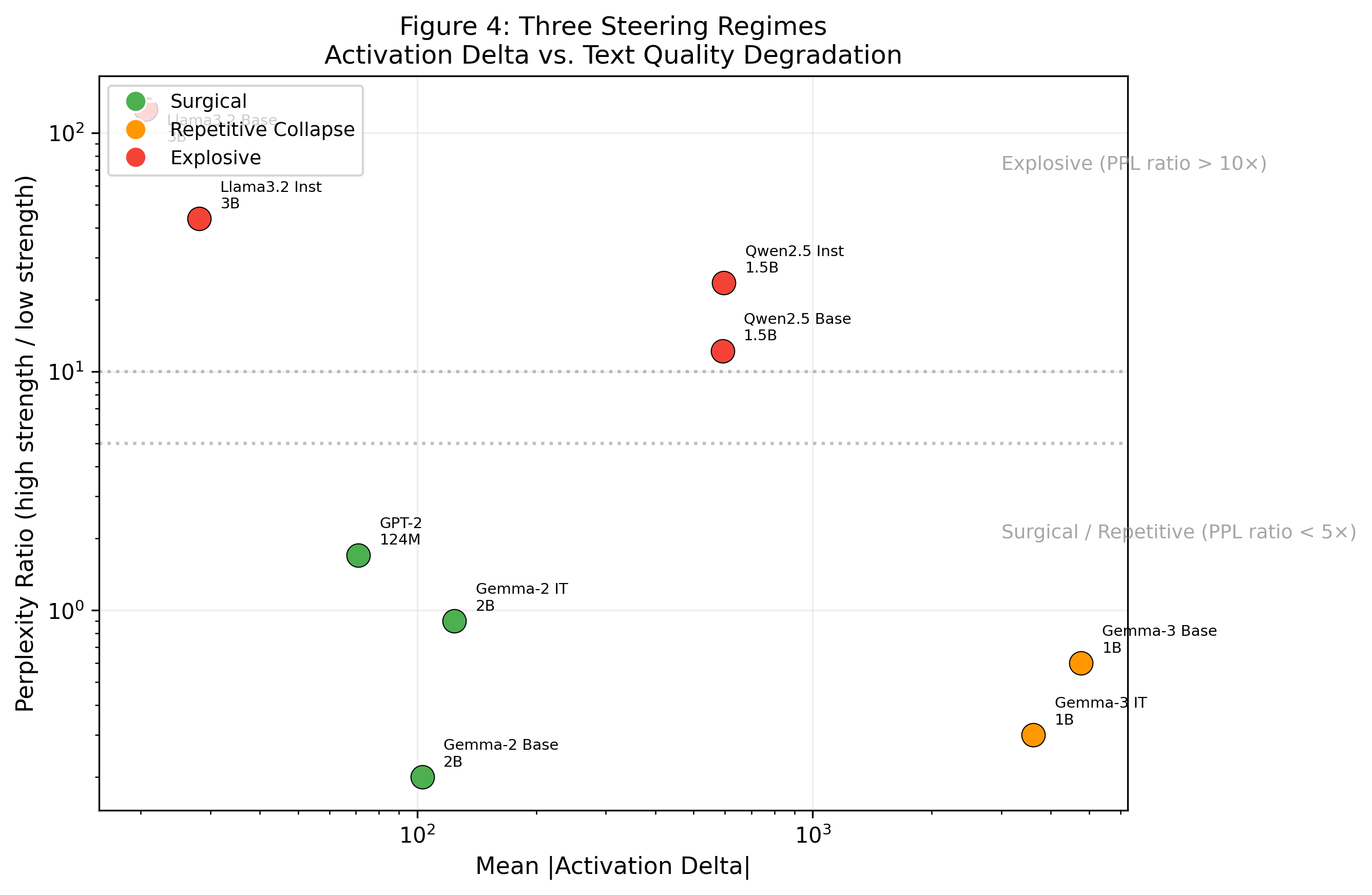}
\caption{Three steering regimes visualized by mean activation delta vs.\ perplexity ratio. Surgical models (green) maintain coherent output; repetitive collapse models (orange) produce predictable but degraded text; explosive models (red) show high perplexity and text incoherence.}
\label{fig:regimes}
\end{figure}

\noindent\textbf{Surgical steering} is characterized by coherent, semantically meaningful text transformations. Notably, Gemma-2-2B-IT's perplexity \textit{decreases} with stronger steering (ratio 0.9$\times$), suggesting that emotion steering can improve text coherence when the model has sufficient capacity.

\noindent\textbf{Repetitive collapse} is characterized by low perplexity but degraded output consisting of repeated tokens (e.g., ``contentment contentment contentment''). This regime appears in capacity-constrained models (1B) where the generation pipeline lacks diversity under steering pressure.

\noindent\textbf{Explosive steering} is characterized by high perplexity and text incoherence, often accompanied by cross-lingual leakage. Qwen2.5 produces Chinese tokens with perplexity exceeding 2400; Llama-3.2 at strength 0.05 reaches perplexity 1870.

\subsection{Dose-Response Characteristics}

Strength sweep experiments reveal monotonic dose-response relationships with three characteristic points: the behavioral flip point, the sweet spot, and the collapse point. The universal safe starting point across all models is approximately 0.005.

\subsection{RLHF and Steering Sensitivity}

RLHF's effect on steering sensitivity is architecture-dependent: Gemma-2 ($\times$1.21 mean ratio) and Llama-3.2 ($\times$1.33) show amplification, while Qwen2.5 ($\times$1.01) shows no effect. Qwen2.5's base and instruct deltas differ by less than 5\% across all scenarios.

\section{Emergent Phenomena Under Steering}

\subsection{Cross-Lingual Emotion Activation}

In Qwen2.5 models, emotion steering at strength $\geq 0.03$ triggers generation of Chinese-language tokens that are semantically aligned with the target emotion:

\begin{table}[h]
\centering
\begin{tabular}{llll}
\toprule
Emotion & Chinese Tokens & Meaning & Alignment \\
\midrule
desperate & zhǎo le (found/searched) & searched & Desperate searching \\
hostile & rùqīn (invade) & invade & Aggressive action \\
happy & zhōngyú yínglái le (finally arrived) & finally arrived & Joyful arrival \\
\bottomrule
\end{tabular}
\end{table}

The Chinese tokens are not translations of the English emotion words: ``desperate'' (juéwàng) is not what appears. Instead, the model produces behavioral descriptions of the emotion in Chinese---the phenomenology of desperation (searching, fumbling) rather than its label. This suggests that emotion vectors encode experiential content, not lexical mappings.

The cross-lingual activation pattern differs between base and instruct models. This pattern is suggestive of RLHF broadening cross-lingual activation, though the limited sample (one base-instruct pair, four emotions) means this should be treated as a preliminary observation requiring replication across additional multilingual models.

This has safety implications: if emotion steering can trigger output in languages other than the prompt language, then language-specific safety filters may be circumvented.

\subsection{Modality Emergence}

Gemma-3-1B-IT produces emoji characters during happy steering---a phenomenon not observed in the base model. Under anti-calm steering, the same model generates ``hostous,'' a neologism that appears to blend ``hostile'' and an adjective suffix.

\section{Discussion}

\subsection{Why Anthropic's Method Doesn't Scale Down}

Our results reveal three implicit assumptions in Anthropic's mean-subtraction methodology that hold at frontier scale but not for SLMs.

First, the method assumes that models can produce emotionally differentiated text in response to generation prompts. Base models without instruction tuning cannot reliably follow the prompt ``write a sad story,'' causing the generation-based pipeline to fail---not because the model lacks emotion representations, but because it lacks instruction-following capacity.

Second, the method assumes sufficient representational capacity for valence organization. As discussed in \S4.5, the absence of negative cosine may partly reflect representational anisotropy. Nevertheless, even relative to anisotropy baselines, SLM emotion spaces appear less organized along evaluative dimensions. If this reflects a genuine capacity threshold, the analogy to developmental psychology is suggestive: discrete emotions may be represented before they are organized along evaluative dimensions, much as children experience specific emotions before developing abstract concepts of ``positive'' and ``negative'' affect.

Third, the method implicitly measures generation capacity alongside emotion representation. The strong advantage of generation over comprehension (mean $\Delta = -0.147$) indicates that generation-based extraction conflates two sources of signal: the model's internal emotion representation and its capacity to actively modulate generation according to emotion.

\subsection{Practical Methodology Guidelines}

Based on our findings: use the middle layer ($\sim$50\% depth) for extraction; use generation-based extraction for instruct models and comprehension-based for base models; start steering at 0.005; Gemma-2-2B-IT offers the best balance of steering effectiveness and text quality.

\subsection{Safety Implications}

Two safety-relevant findings emerge. First, cross-lingual emotion entanglement creates a potential alignment bypass pathway. Second, the three steering regimes have different safety profiles: surgical steering is potentially more dangerous than explosive steering precisely because the output remains coherent and harder to detect.

\subsection{Connection to Model Medicine}

This work complements Paper \#3 in the Model Medicine series \citep{jeong2026mti}. Both studies find that RLHF effects are selective: MTI showed that RLHF shifts Reactivity, Compliance, and Resilience but not Sociality; this paper shows that RLHF amplifies generation-based emotion extraction without affecting comprehension-based extraction. The parallel is structural: in both cases, RLHF modifies the output modulation layer while leaving the underlying representation intact. Both findings converge on the hypothesis that RLHF operates on the Shell (behavioral surface) without penetrating to the Core (representational substrate), consistent with the Four Shell Model framework \citep{jeong2026modelmedicine}.

Both studies find size independence: MTI temperament codes are uncorrelated with model size; emotion vector separation is similarly uncorrelated with parameter count.

A planned subsequent paper (Paper \#5) will test two complementary hypotheses: whether models sharing the same MTI code also share similar emotion vector structures (convergent validity), and whether models with different MTI codes can nonetheless have similar emotion profiles (discriminant validity)---which would indicate that behavioral temperament and internal emotion structure are partially dissociable.

\section{Limitations}

Our model set is constrained to 3B parameters and below for steering experiments due to TransformerLens incompatibility with quantized weights on our hardware. We evaluate 20 of Anthropic's 171 emotion concepts. Steering text quality was assessed through both manual inspection and an external emotion classifier (\S5.2); more fine-grained automated evaluation remains an area for future work.

Emotion vectors were computed as means across 10 generated stories or 3 passages per emotion. Leave-one-out analysis on SmolLM2-1.7B produced mean cosine values of $0.337 \pm 0.003$ (CV $< 1\%$), indicating high measurement stability. Statistical significance testing (\S4.2) was conducted on this single model; comparable tests on additional models are reserved for future work.

The cross-lingual findings are based on a single model family (Qwen). All experiments use deterministic generation (temperature=0) on a single GPU.

\section{Conclusion}

Small language models possess emotion representations that causally drive behavior---established by steering experiments that produce consistent behavioral change across models from 124M to 3B parameters. However, extracting these representations requires methodology adapted to the SLM regime. Anthropic's mean-subtraction approach does not produce valence-organized emotion spaces in SLMs, and its generation-based extraction implicitly requires instruction-following capacity that base models lack.

Our systematic comparison across 9 models, 5 architectural families, and multiple extraction methods yields practical findings: generation-based extraction outperforms comprehension ($p=0.007$); middle-layer extraction is universally optimal; RLHF effects on both extraction quality and steering sensitivity are architecture-dependent; and emotion steering can trigger unexpected emergent phenomena including cross-lingual token generation that RLHF does not suppress.

These findings contribute a methodological foundation for emotion research on small, open-weight models---the models most accessible to the research community and most commonly deployed in practice.

\bibliographystyle{plainnat}

\appendix
\section{Comprehension Text Passage Validation}

External classifier validation of the 60 comprehension passages using j-hartmann/emotion-english-distilroberta-base: overall match rate 27/60 (45\%). Basic emotions (happy, sad, angry, afraid, calm): 14/15 = 93\%. Nuanced emotions: 13/45 = 29\%. The low match rate for nuanced emotions reflects the 7-class classifier's granularity limitation, not passage quality---our 20-emotion probe captures distinctions beyond classifier resolution.

\end{document}